\documentclass[10pt,twocolumn,letterpaper]{article}

\usepackage{cvpr}              

\usepackage[dvipsnames]{xcolor}

\usepackage{wrapfig}

\definecolor{cvprblue}{rgb}{0.21,0.49,0.74}
\usepackage[pagebackref,breaklinks,colorlinks,citecolor=cvprblue]{hyperref}

\usepackage[symbol]{footmisc}

\def\paperID{15266} 
\def\confName{CVPR}
\def\confYear{2024}

\title{Pix2Repair: Implicit Shape Restoration from Images}

\author{
Xinchao Song*, Nikolas Lamb*, Sean Banerjee, Natasha Kholgade Banerjee\\
Clarkson University, Potsdam, New York, USA \\
\texttt{\{xisong, lambne, sbanerje, nbanerje\}@clarkson.edu}
}

\begin{document}

\twocolumn[{%
\renewcommand\twocolumn[1][]{#1}%
\maketitle
\begin{center}
    \centering
    \captionsetup{type=figure}
    \includegraphics[width=\linewidth]{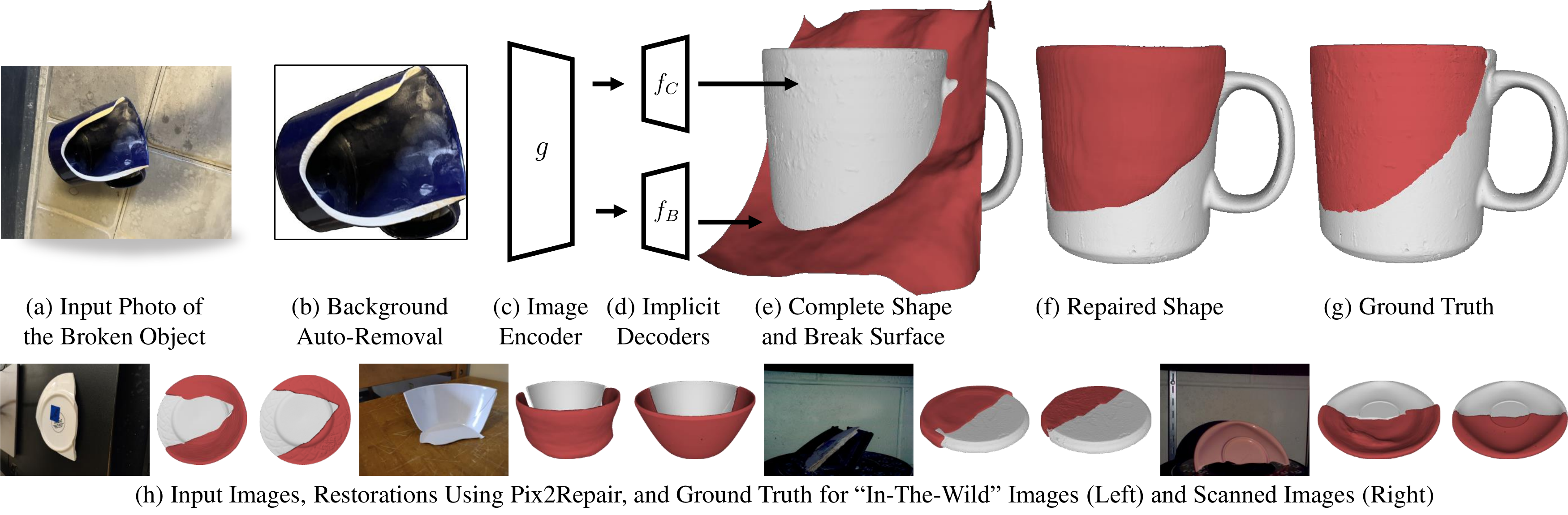}
    \captionof{figure}{We present Pix2Repair, an automated shape repair approach. Pix2Repair takes an image of a fractured object as input, automatically removes the background, and predicts a complete shape and break surface, which combine to generate a restoration shape.}
    \label{fig:teaser}
\end{center}%
}]

\footnote[0]{*These authors contributed equally to this paper.}

\begin{abstract}

We present Pix2Repair, an automated shape repair approach that generates restoration shapes from images to repair fractured objects. Prior repair approaches require a high-resolution watertight 3D mesh of the fractured object as input. Input 3D meshes must be obtained using expensive 3D scanners, and scanned meshes require manual cleanup, limiting accessibility and scalability. Pix2Repair takes an image of the fractured object as input and automatically generates a 3D printable restoration shape. We contribute a novel shape function that deconstructs a latent code representing the fractured object into a complete shape and a break surface. We also introduce \textit{Fantastic Breaks Imaged}, the first large-scale dataset of 11,653 real-world images of fractured objects for training and evaluating image-based shape repair approaches. Our dataset contains images of objects from \textit{Fantastic Breaks}, complete with rich annotations. We show restorations for real fractures from our dataset, and for synthetic fractures from the Geometric Breaks and Breaking Bad datasets. Our approach outperforms shape completion approaches adapted for shape repair in terms of chamfer distance, normal consistency, and percent restorations generated.

\end{abstract}

\section{Introduction}
Many people know the frustration of trying to repair a treasured glass, vase, or mug that has been accidentally fractured, e.g., Figure~\ref{fig:teaser}(a). Sometimes fractured objects can be repaired via reassembly. However, in many cases parts are destroyed or lost during the fracturing process leaving gaps in the repaired object. Automated shape repair approaches seek to fill this gap by automatically generating restoration shapes for fractured objects. In addition to preserving heirlooms, object repair algorithms have the potential to reduce household waste, restore cultural heritage objects, and automate industrial recycling. A recent body of work has emerged to automate the repair of fractured objects by predicting 3D printable restoration shapes using machine learning~\cite{hermoza20183d, lamb2022mendnet, lamb2022deepmend, lamb2022deepjoin}.  However, existing approaches operate in low-resolution voxel space~\cite{hermoza20183d}, or require high resolution 3D scans of the fractured object~\cite{lamb2022mendnet, lamb2022deepmend, lamb2022deepjoin} which must be obtained using expensive scanners, e.g. the \$2,000 Einscan-SP, making them inaccessible to consumers.

We present Pix2Repair, an approach to automatically generate repair parts for fractured objects from images. Our approach builds on prior approaches that perform shape completion from images using implicit representations~\cite{murez2020atlas, lin2020sdf, mescheder2019occupancy, chen2019learning, wang2022mdisn, li2021d2im, xu2020ladybird, hui2022neural}. However, existing shape completion approaches generate a complete shape from a partial observation, e.g., an image that observes the complete shape from a fixed perspective. Different from shape completion, our approach generates a restoration shape from an image of a fractured object, such as the image shown in Figure~\ref{fig:teaser}(a). Generating restoration shapes is a harder task than generating complete shapes as restorations are often much smaller than complete shapes, as is the case for the restoration shape in Figure~\ref{fig:teaser}(f). Restorations also contain high-frequency geometry at the point of fracture 
that is difficult to predict directly. 

To overcome the challenge of representing shapes with small size and high-frequency surface detail, we compute the repair shape as the Boolean intersection of a predicted complete shape and break shape, where the break shape is given by a break surface, shown in Figures~\ref{fig:teaser}(e). Our approach draws inspiration from existing shape repair approaches~\cite{lamb2022deepmend, lamb2022deepjoin}, which similarly represent the restoration shape as a Boolean intersection. However, our approach learns to predict a complete shape and break surface from a single shape code representing the fractured object, in contrast to prior approaches that obtain two separate complete and break codes by performing iterative optimization during inference. Learning a mapping from the fractured shape code to the complete shape and break surface allows our network to predict restoration shapes directly from color images of the fractured shape. Accepting images as input makes Pix2Repair more accessible to consumers and faster during inference than prior approaches. 

As no publicly available datasets of real images of fractured objects exist, we also introduce \textit{Fantastic Breaks Imaged}, the \textit{first large-scale dataset of real-world images of fractured objects} as a benchmark dataset for image-based repair approaches. Real fractured objects depicted in our dataset have corresponding high-resolution 3D scans in \textit{Fantastic Breaks} dataset~\cite{lamb2023fantastic} to enable training or evaluation of image-based repair algorithms. 

We make the following contributions:
\begin{enumerate}
    \item We present the first approach to generate implicit restoration shapes from color images. 
    \item We introduce a novel network architecture that learns a mapping directly from a fractured shape code to a complete shape and break surface with after performing image background auto-removal.
    \item We contribute a dataset of 11,653 images with rich annotations of damaged objects from \textit{Fantastic Breaks}, including 9,528 images taken “in-the-wild” and 2,109 scanned images.
    \item We demonstrate the effectiveness of our approach by repairing real and synthetically fractured objects.
\end{enumerate}

We perform training using real fractures from \textit{Fantastic Breaks Imaged} and synthetic fractures from the Breaking Bad~\cite{sellan2022breaking} and Geometric Breaks~\cite{lamb2022deepjoin} datasets. We evaluate our approach on \textit{Fantastic Breaks Imaged}. We outperform two shape completion approaches that we adapt for shape repair in terms of chamfer distance, normal consistency, and percent of restorations generated. We show that use of real images of fractured objects from our dataset for training repair approaches results in more accurate reconstructions.

The code and dataset will be available at the time of the official publication of this paper.

\section{Related Work}

\subsection{Shape Repair}

Prior work in shape repair has been mostly manual, with approaches only repairing a single object, e.g. a terracotta artifact~\cite{scopigno20113d}, a silver crown~\cite{seixas2018use}, a fruit bowl~\cite{antlej2011combining}, a dinosaur vertebrae~\cite{schilling2014reviving}, a rhinoplasty implant~\cite{bekisz2019house}, or a dental prosthesis~\cite{rengier20103d}. Manual approaches require that the user generate all or part of the restoration using 3D design software, which is out of the scope of the average consumer. Some automated approaches generate repairs by exploiting object symmetry, i.e. using symmetric reflection and subtraction~\cite{sipiran2018completion, li2014symmetry, mavridis2015object}, or performing shape subtraction with a high-resolution oracle~\cite{lamb2019automated}. However, these approaches fail to repair asymmetric objects~\cite{sipiran2018completion, li2014symmetry, mavridis2015object}, or cannot repair objects without a corresponding complete 3D scan~\cite{lamb2019automated}. 

Early machine learning based repair approaches operate in $32^3$ voxel space~\cite{hermoza20183d}, generating coarse repair parts that cannot accurately represent the complex fractured surface. More recent work in shape repair~\cite{lamb2022deepmend, lamb2022mendnet, lamb2022deepjoin} uses the autodecoder framework introduced by DeepSDF~\cite{park2019deepsdf}. These approaches generate restoration shapes by learning a mapping from fractured shapes to restoration shapes~\cite{lamb2022mendnet}, or learn occupancy~\cite{lamb2022deepmend} or combined signed distance, occupancy, and normal field functions~\cite{lamb2022deepjoin} for a complete shape and break surface that combine to generate the restoration shape. Similar to prior approaches~\cite{lamb2022deepjoin, lamb2022deepmend}, our approach represents the restoration shape as a Boolean intersection. However, unlike prior work our shape functions take a latent code corresponding to the fractured shape as input, instead of two separate codes for the complete shape and break surface. By inputting a single fractured shape code, our approach learns to map fractured shape observations to complete shapes and break surfaces, allowing us to perform inference directly from color images of the fractured shape. 

\subsection{Shape Completion}

\begin{figure}
\centering
\includegraphics[width=\linewidth]{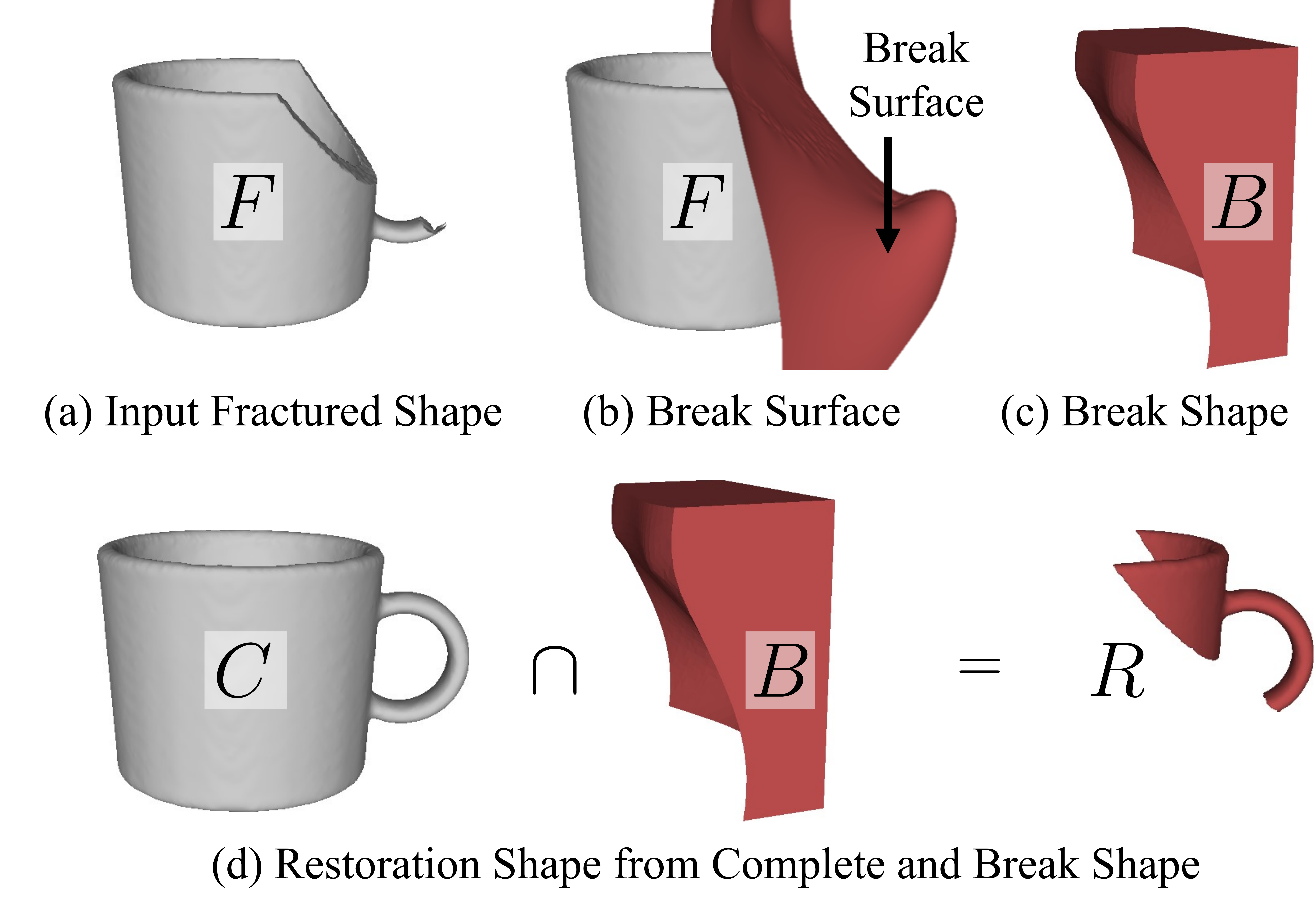}
\caption{(a) Given a fractured shape $F$, (b) the break surface intersects with $F$ at the point of fracture, (c) and the break shape $B$ is the point set on the restoration side of the break surface. (d) The union of the complete shape $C$ and break shape $B$ gives the restoration shape $R$.}
\label{fig:setrel}
\end{figure}

A related but still disparate field of study from shape repair is shape completion, which focuses on generating a complete shape from a color image. Recent work in shape completion from images can be categorized according to the method of shape representation. Some approaches perform completion on point clouds~\cite{ping2022visual, yu2022part, wu2020pq, mandikal20183d}, which are compact and are easy to obtain, e.g., using depth cameras. However, as point clouds lack orientable surfaces they cannot be used to generate 3D-printable restoration parts. Though mesh based approaches~\cite{shi2021geometric, gkioxari2019mesh, wang20193dn, groueix2018papier, wu2018learning, wang2018pixel2mesh} produce closed, orientable surfaces that are necessary to represent physically realizable restoration shapes, they struggle to represent shapes of arbitrary topology or overly complex meshes~\cite{mescheder2019occupancy}. Voxel approaches have been extensively studied for shape representation, including large-scale scene reconstruction~\cite{popov2020corenet}, single and multi-view object completion~\cite{choy20163d, xie2019pix2vox, xie2020pix2vox++}, and completion from 2.5D images~\cite{wu2017marrnet}. Recent work has improved the efficiency of voxel approaches using, e.g., sparse convolutions~\cite{dai2020sg, yi2021complete} and hierarchical models~\cite{dai2020sg, wu2022voxurf}. However, despite advancements in efficiency, voxel approaches still discretize the output space, limiting their ability to represent the high-frequency geometry present at the fractured region of restoration shapes.

A large body of recent work has focused on learning to represent shapes implicitly using functions~\cite{murez2020atlas, lin2020sdf, mescheder2019occupancy, chen2019learning, wang2022mdisn, li2021d2im, xu2020ladybird, hui2022neural}. Shape functions can reconstruct shapes at arbitrarily high resolutions, making them well suited for restoration shape prediction. Occupancy networks~\cite{mescheder2019occupancy} was one of the first approaches to explore using the parameterized occupancy function to perform single-view reconstruction. Later approaches would propose modifications to shape functions such as learning multi-scale implicit fields~\cite{wang2022mdisn}, disentangling object geometry and texture~\cite{li2021d2im}, imposing symmetry constraints~\cite{xu2020ladybird}, or learning to deform shape templates~\cite{hui2022neural}. Unlike prior shape completion approaches, our approach learns to predict a restoration shape as the Boolean intersection of two other shape functions, i.e. the complete shape and break surface. In Section~\ref{sec:res} we demonstrate that by representing the restoration shape as a Boolean intersection, our approach outperforms shape completion approaches that are adapted to perform shape repair in terms of the chamfer distance, normal consistency, and percent of restorations generated.

\subsection{Fractured Object Datasets}

Existing datasets with fracture-based damage to objects may be broadly divided into two categories\textemdash{}synthetic and real-world. Synthetic fractured object datasets, introduced to broaden the scale of data for tasks such as learning-based repair and assembly include Breaking Bad~\cite{sellan2022breaking}, Geometric Breaks~\cite{lamb2022deepjoin}, and the Neural Shape Mating (NSM) dataset~\cite{chen2022neural}. The NSM dataset contains 1,246 damaged objects generated using simple planar, sine, parabolic, square, and pulse cut functions. Geometric Breaks contains 25,250 fractures generated through the subtraction of simple 3D shapes, with perturbation of the 3D shape to represent the roughness seen in ceramics. Both the NSM dataset and Geometric Breaks represent fracture using a heuristics approach. Though 3D shape datasets can be used in image-based repair by generating renders, due to the limited diversity of synthetic datasets using exclusively rendered images in training results in poor shape repair results when tested on images of real fractured objects. 

To the best of our knowledge, only two real-world publicly available datasets of fractured objects exist\textemdash{}the Hampson Museum cultural heritage dataset~\cite{payne2009designing} and \textit{Fantastic Breaks}~\cite{lamb2023fantastic}. Apart from these datasets, all prior work on repair or reassembly has been done for a small set of objects, e.g., 3~\cite{brown2008system,funkhouser2011learning}, 5~\cite{hong2019potsac}, 7~\cite{huang2006reassembling}, and 22~\cite{lamb2019automated} objects. Though two approaches obtain images and depth scans~\cite{brown2008system,funkhouser2011learning}, they provide only 3 objects. The Hampson Museum dataset contains 3D scans of 138 cultural heritage objects with varying levels of damage. The dataset lacks paired complete scans, preventing it from being used to train repair approaches. \textit{Fantastic Breaks} contains 150 paired broken object 3D scans and complete counterpart 3D scans. Similar to the synthetic datasets, both these datasets are 3D shape-oriented datasets and lack images. Training for real shape repair on renders is likely to suffer from bias in synthetic-to-real methods~\cite{gondal2019transfer}.

\section{Representing Restoration Shapes}

The goal of our approach is to generate a restoration shape from an image of the fractured shape, such that the restoration shape can be used to repair the fractured shape. We represent shapes as open point sets. We represent the occupancy of a point $\mathbf{x}$ in a given shape $S \in \{C, F, R\}$ as $o_{S}(\mathbf{x})$, such that $o_{S}(\mathbf{x})=1$ if $\mathbf{x} \in S$ and $o_{S}(\mathbf{x})=0$ otherwise, where $C$, $F$, and $R$ refer to the complete, fractured, and restoration shapes respectively. We define the break surface, shown in Figure~\ref{fig:setrel}(b), as a 2D manifold that intersects the fractured and restoration shapes at the point of fracture. We define the break shape $B$, shown in Figure~\ref{fig:setrel}(c), as the set of points on the restoration side of the break surface, i.e. $o_{B}(\mathbf{x})=1$ if $\mathbf{x}$ is on the restoration side of the break surface and $o_{B}(\mathbf{x})=0$ if $\mathbf{x}$ is on the fractured side of the break surface. Though the break shape extends infinitely out from the fractured shape, in practice we limit it to the unit cube. We use an open set representation to prevent points from belonging to $F$ and $R$ simultaneously. 

Using the definition for $B$, the restoration shape can be represented in terms of the complete shape and break shape as $R = C \cap B$. We show an example restoration shape obtained through the intersection of a complete and break shape in Figure~\ref{fig:setrel}(d). We follow the approach of Lamb et al.~\cite{lamb2022deepmend} to obtain $R$ using the occupancy function definitions for $C$ and $B$. We obtain the occupancy equation for $R$ in terms of a given point $\mathbf{x}$ as $o_{R}(\mathbf{x}) = o_{C}(\mathbf{x}) \land o_{B}(\mathbf{x})$, where $\land$ represents the logical \texttt{and} operator. To enable $o_{R}(\mathbf{x})$ to work with continuous values, i.e. to be used in a loss function, we relax the equation for $o_{R}(\mathbf{x})$ using using the product T-norm~\cite{gupta1991theory}, such that 
\begin{align}
    \label{eq:or}
    o_{R}(\mathbf{x}) & = o_{C}(\mathbf{x}) o_{B}(\mathbf{x}). 
\end{align}
Equation~\ref{eq:or} allows us to obtain the occupancy for $R$ in terms of the predicted occupancy for $C$ and $B$.

\section{Pix2Repair}

We learn two functions, $f_C$ and $f_B$, to represent occupancy in the complete shape and occupancy in the break shape respectively. During training, the function $f_C$ is tasked with learning a one-to-many mapping from the fractured shape observation to the complete shape. The function $f_B$ is tasked with learning a one-to-one mapping from the fractured shape observation to the break shape.   Each function takes point $\mathbf{x} \in \mathbb{R}^3$ as input and is  conditioned on a latent code corresponding to an observation of the fractured shape $\mathbf{z}$, such that $o_{C}(\mathbf{x}) = f_{C}(\mathbf{z}, \mathbf{x})$ and $o_{B}(\mathbf{x}) = f_{B}(\mathbf{z}, \mathbf{x})$. During training and testing we obtain $\mathbf{z}$ by passing an input image through an image encoder $g$ to obtain a fractured shape latent code. We compute occupancy in the restoration shape by substituting the predicted values into Equation~\ref{eq:or}, i.e. as $o_{R}(\mathbf{x}) = f_{B}(\mathbf{z}, \mathbf{x})f_{C}(\mathbf{z}, \mathbf{x})$. 


We train our network on complete, break, and restoration shape tuples $(C, B, R)$. We use $\hat{o}_{C}(\mathbf{x})$, $\hat{o}_{B}(\mathbf{x})$, and $\hat{o}_{R}(\mathbf{x})$ to denote the ground truth occupancy values for shapes $C$, $B$, and $R$ respectively at a point $\mathbf{x}$. During training, for each shape tuple we optimize with respect to the loss 
\begin{align}
    \mathcal{L} = \sum_{\mathbf{x} \in \mathcal{X}} \left( \mathcal{L}_{C} + \lambda_{B}\mathcal{L}_{B} + \lambda_{R}\mathcal{L}_{R} \right),
\end{align}
where
\begin{align}
    \label{eq:lc}
    \mathcal{L}_{C} & = BCE \left( o_{C}(\mathbf{x}), \hat{o}_{C}(\mathbf{x}) \right), \\
    \label{eq:lb}
    \mathcal{L}_{B} & = BCE \left( o_{B}(\mathbf{x}), \hat{o}_{B}(\mathbf{x}) \right), \ \textrm{and}  \\
    \label{eq:lr}
    \mathcal{L}_{R} & = BCE \left( o_{R}(\mathbf{x}), \hat{o}_{R}(\mathbf{x}) \right).
\end{align}
In Equations~\eqref{eq:lb},~\eqref{eq:lr},~and~\eqref{eq:lr}, $BCE$ is the binary cross-entropy loss, and $\mathcal{X}$ is a set of probing points. The losses $\mathcal{L}_{C}$, $\mathcal{L}_{B}$, and $\mathcal{L}_{R}$ encourage the predicted values for the complete, break, and restoration shapes to converge to the ground truth values during training. $\lambda_{B}$ and $\lambda_{R}$ are coefficients that allow us to balance the relative importance of the complete, break, and restoration shape loss.

During training we optimize with respect to the parameters of $g$, $f_{C}$, and $f_{B}$. We use $\lambda_{B} = 1.0$ and $\lambda_{R} = 1.0$. To perform optimization we use the Adam~\cite{kingma2014adam} optimizer with a learning rate of $lr=2e-5$. For all experiments we train our approach for 10 epochs. 

Unless otherwise specified we use ResNet34 as our image encoder $g$. We concatenate ResNet34 with a fully connected layer with 512 output nodes to obtain a 512 dimensional fractured shape code $\mathbf{z}$. We represent the occupancy functions $f_{C}$ and $f_{B}$ using two neural networks with 6 ResNet blocks and batch normalization~\cite{ioffe2015batch}. Each ResNet block has 256 nodes with leaky ReLU activation functions. The latent code and points are input to $f_C$ and $f_B$ separately using two fully connected layers with 256 output nodes and added together after the first layer. The final layer has a single output node with a sigmoid activation function. 

To obtain sample points for training, for each training shape tuple we pre-compute $n$ sample points uniformly in a cube with side lengths of $1.1$ units surrounding the shapes. We also pre-compute $n$ sample points on the surface of each shape in the tuple, and offset the surface points randomly according to a Gaussian distribution with $\sigma=0.01$. We combine the uniform and surface sampled points into a single set $\mathcal{P}$. We use $n=50,000$. On each training iteration we randomly select a subset $\mathcal{X} \subset \mathcal{P}$ of $m$ sample points from the pre-computed uniform and surface sampled points. We select the points $\mathcal{X}$ such that, for each shape $S \in \{C, B, R\}$, at least $\frac{m}{6}$ points lie inside of $S$ and at least $\frac{m}{6}$ points lie outside of $S$.  During training, we use $m=2,048$ sample points. After training our network, to obtain the restoration mesh we query our network with a grid of $k^3$ points and extract the $0.5$ level isosurface using marching cubes~\cite{lorensen1987marching}. For evaluation we use $k=128$. For the visualizations shown in the paper we use $k=256$.

\begin{figure}[t]
    \centering
    \includegraphics[width=\linewidth]{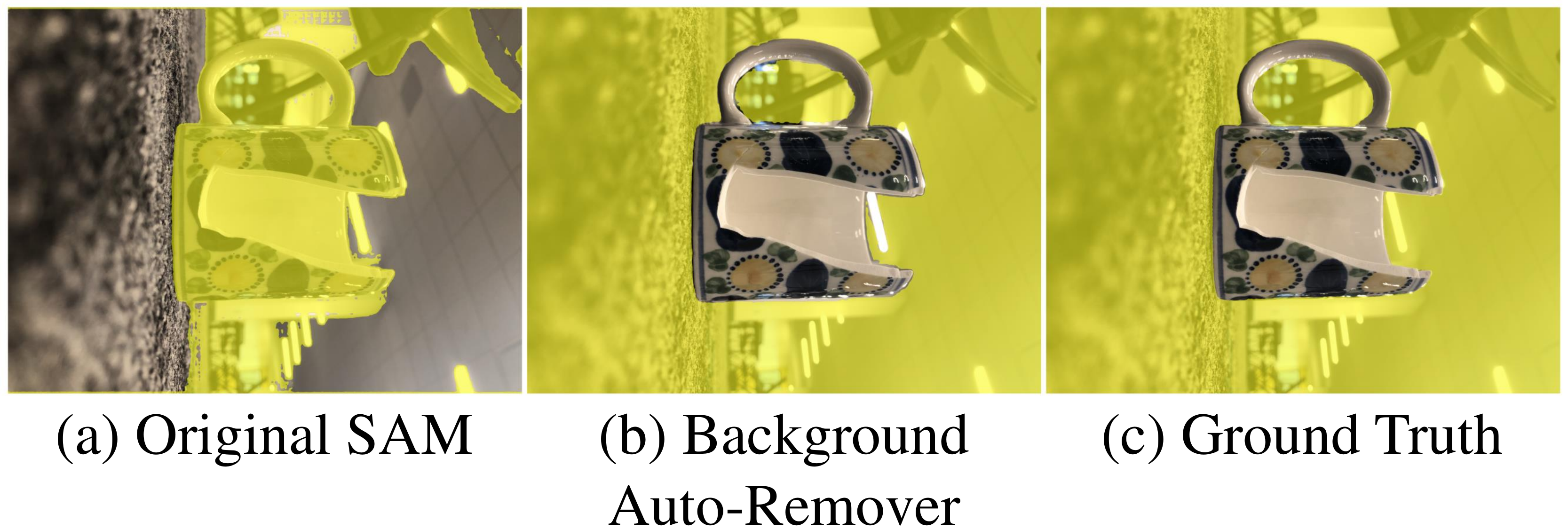}
    \caption{Background removal of the original SAM, our background auto-remover, and the ground truth.}
    \label{fig:remover}
\end{figure}

We input images of fractured objects with backgrounds removed to our network. We separately train a background auto-remover to automatically detect the fractured object and remove the background using the images and ground truth masks from our \textit{Fantastic Breaks Imaged} dataset. We use Segment Anything Model (SAM)~\cite{kirillov2023segment} as the base model and fine-tune its mask decoder to generate favorable masks based on the approach proposed by MedSAM~\cite{Ma2023SegmentAI}. A bounding box of the same size as the input image is used as the prompt for the decoder. To further remove the noise, only the largest connected component in the generated mask is returned as the final mask. We use Dice Similarity Coefficient (DSC) to evaluate the removal performance. Our auto-remover achieves an average DSC score of 0.9608, while the original SAM achieves a score of 0.0001. An example using our background remover is shown in the Figure \ref{fig:remover}.

\begin{figure}[t]
    \centering
    \includegraphics[width=\linewidth]{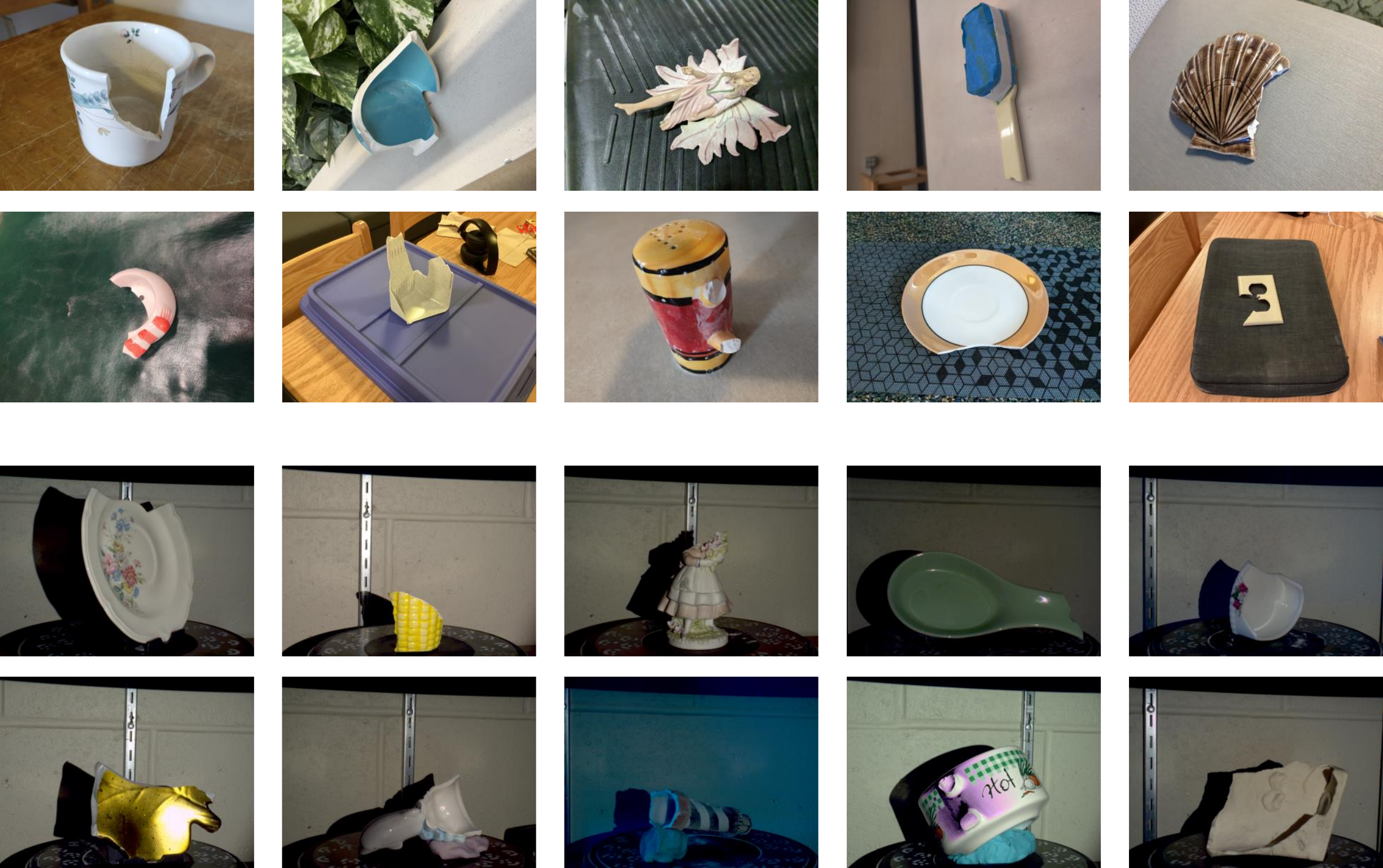}
    \caption{Example ITW photographs (top) and scanned images (bottom) from our dataset.}
    \label{fig:ex}
\end{figure}

\section{Fantastic Breaks Imaged Dataset}\label{sec:dataset}

Our dataset contains 11,653 images of damaged objects from \textit{Fantastic Breaks}, allowing shape repair approaches to use 3D models from \textit{Fantastic Breaks} for training and evaluation. 9,528 images are taken ``in-the-wild'', i.e., photographed in a variety of residential, office, and classroom environments, and 2,109 scanned images from the imaging process used to generate 3D scans for Fantastic Break with six degree-of-freedom (6DOF) pose annotations, colored depth scans with normals. All images come with masks.


\begin{figure*}[t]
    \centering
    \includegraphics[width=\linewidth]{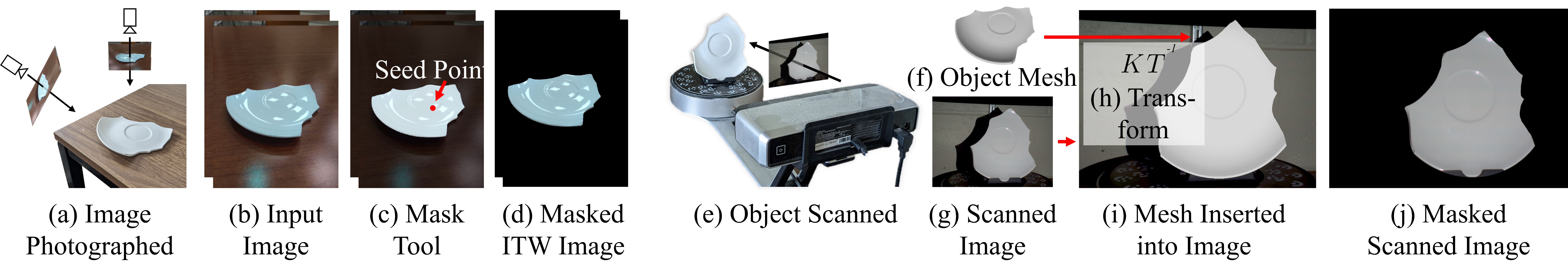}
    \caption{Data collection. For ITW images, we (a-b) photograph objects, and (c) use SAM to (d) obtain masked images. For scanned images, we use the model and .xml files obtained from Einscan scanner 3D scan process in \textit{Fantastic Breaks} (scanning process shown in (e)), to (f-g) obtain a mesh and image, and use the transform in (h) to (i) insert the mesh into the image and (j) obtain the masked image.}
    \label{fig:overview}
\end{figure*}

\paragraph{``In-The-Wild'' (ITW) Images.} We collect 9,528 images of fractured objects belonging to \textit{Fantastic Breaks}~\cite{lamb2023fantastic} using consumer grade phones and cameras. We recruited 14 experimenters to obtain photographs. Experimenters were instructed to capture objects as if they were inputting the images to a black-box automated repair algorithm that requires knowledge of the fractured region. Each object was photographed by at least two experimenters, from multiple viewpoints in at least three locations, as shown in Figure~\ref{fig:overview}(a). Images depict objects in their entirety, i.e., objects are not partially obscured or cropped at the image boundary. Photographs were taken in diverse locations and against a variety of backgrounds within university premises. We show example images at the top of Figure~\ref{fig:ex}. 

We use Segment Anything Model (SAM)~\cite{kirillov2023segment} to generate masks that identify the object in each image. Two experimenters manually selected a seed point as input to SAM, and selected a segmentation level (out of three possible levels) that best segmented the object, as shown in Figure~\ref{fig:overview}(c). We built a tool for experimenters to select a seed point, processes images in a background thread, and displays results after processing, enabling rapid annotation. Experimenters favored over-segmentation in the case of imperfect segmentations. Figure~\ref{fig:overview}(d) shows a masked image example.

\paragraph{Scanned Images.} We obtain 2,109 images by parsing scan files of the objects provided by Fantastic Breaks~\cite{lamb2023fantastic} through scanning the object using the Einscan SP 3D scanner. The scan process is shown in Figure~\ref{fig:overview}(e). We collected images from the scan files obtained by the authors of \textit{Fantastic Breaks}. During scanning, the scanner software produces a single \texttt{.xml} project file that contains pointers to the color images and depth scans, as well as the scanner intrinsic matrix. Color images are saved as \texttt{.bmp} files. Depth scans are saved as binary \texttt{.rge} files, which also store the camera frame, distance from the camera to the scanner turntable, depth scan points, colors, and normals, and the transformation matrix $T$ that aligns the depth scan to the final high-resolution 3D model. We implement a parser for the \texttt{.rge} files that enables extraction of the depth scan points, colors, and normals, and the object pose in the image $T$. We used our parser to extract colored point clouds with normals that correspond to the depth scans obtained by the scanner. We show example scanned images at the bottom of Figure~\ref{fig:ex}.  We obtained the mask of the object in each \texttt{.bmp} image using the transformation matrix $T$ stored in the \texttt{.rge} file. We invert $T$ to obtain a transform that inserts the 3D model in the \texttt{.bmp} image, $T^{-1}$. We apply the transform $T^{-1}$ to the 3D model, and project the transformed model onto the image using the camera intrinsic matrix $K$ to obtain the image mask, as shown in Figure~\ref{fig:overview}(i).

\setlength{\tabcolsep}{1.2pt}
\begin{table}[t]
\centering
\footnotesize
\caption{Per-class image counts for fractured objects from \textit{Fantastic Breaks} (FB), and depth scans, raw scan files, and images from our dataset, \textit{Fantastic Breaks Imaged}.}
\begin{tabular}{l|cccccccccccc|c}
\toprule
 & Mug & Plate & Statue & Bowl & Cup & Jar & Coaster & Box & Misc & Total \\ \hline
Objects (FB) & 30 & 35 & 30 & 17 & 8 & 6 & 6 & 3 & 15 & 150 \\ \hline
Depth & 902 & 889 & 774 & 412 & 193 & 128 & 147 & 90 & 339 & 3,874 \\
Raw Scanned & 902 & 889 & 774 & 412 & 193 & 128 & 147 & 90 & 339 & 3,874 \\ 
Raw Depth & 902 & 889 & 774 & 412 & 193 & 128 & 147 & 90 & 339 & 3,874 \\ \hline
ITW & 1,999 & 2,526 & 1,726 & 1,146 & 511 & 419 & 434 & 120 & 647 & 9,528 \\
Scanned & 491 & 447 & 427 & 215 & 124 & 91 & 72 & 48 & 194 & 2,109 \\ \hline
Total & 2,490 & 2,973 & 2,153 & 1,361 & 635 & 510 & 506 & 168 & 841 & 11,637 \\
\bottomrule
\end{tabular}
\label{tab:class_file_sizes}
\end{table}



\paragraph{Data Validation.} To ensure high-quality and consistent data, one experimenter inspected all images and masks in our dataset. Shape repair approaches require inputs that view the missing region of the object to generate accurate repairs. We discarded images that did not view the missing region of the object. We discarded images containing objects that were more than 10\% occluded by foreground objects. We also discarded images that were blurry, had inaccurate masks, or had color aberrations.

\paragraph{Dataset Analysis.} \textit{Fantastic Breaks} contains 150 objects over 9 classes of commonly damaged household objects, covering mugs, plates, statues, bowls, cups, jars, coaster, boxes, and a miscellaneous class. Objects have ceramic (115 objects), plastic (13 objects), glass (7 objects), plaster (4 objects), wood (3 objects), and other (8 objects) material compositions. We collect scanned images for all 150 objects, and collect ITW images for 148 of the 150 objects, as 2 objects were unobtainable. The number of 3D scanned objects from \textit{Fantastic Breaks} and the number of scanned and ITW images from \textit{Fantastic Breaks Imaged} over each class is shown in Table~\ref{tab:class_file_sizes}. We provide the ITW and scanned images as \texttt{.png} files and depth maps as \texttt{.ply} files. Raw depth scans and raw scanned images before extraction and data validation are also provided as binary \texttt{.rge} and \texttt{.bmp} files.

\section{Experiments}\label{sec:experiments}

We train Pix2Repair on different combinations of \textit{Fantastic Breaks Imaged} and two synthetic datasets, and evaluate it on \textit{Fantastic Breaks Imaged}. As no image-based shape repair approaches exist as baselines for comparison, we adapt two existing shape completion approaches, Occupancy Networks (ONet)~\cite{mescheder2019occupancy} and Pix2Vox++~\cite{xie2020pix2vox++}, to take images as input to perform shape repairs. Since synthetically generated images do not have backgrounds, for a fair comparison we also use directly the masked images from our dataset without background auto-removal, which were generated by applying the ground truth masks.

\subsection{Synthetic Datasets}

We use two publicly available fractured object datasets that contain synthetically fractured objects: Geometric Breaks~\cite{lamb2022deepjoin} and Breaking Bad~\cite{sellan2022breaking}.

Geometric Breaks contains 24,208 objects from ShapeNet~\cite{shapenet2015}, and 1,042 objects from the Google Scanned Objects Dataset (GSO)~\cite{downs2022google}. Each object is fractured between 1 and 10 times. Objects are fractured by subtracting a randomized geometric primitive from each object. We use objects from the jars, bottles, mugs, and GSO subsets of the dataset, totalling 5,252 objects.

Breaking Bad contains 10,474 objects from PartNet~\cite{mo2019partnet} and Thingi10K~\cite{zhou2016thingi10k}. The dataset is divided into everyday objects, artifact objects, and other objects based on application. Fractures are synthesized by computing a set of the most likely fracture modes. Each object is fractured 80 times. We use 405 objects from the everyday subset, totaling 32,400 objects.

All objects from the Geometric Breaks and Breaking Bad datasets are normalized to sit inside a unit cube. Though all objects from Geometric Breaks are oriented upright, some objects from Breaking Bad are not. We manually find and remove objects from Breaking Bad that are oriented perpendicular to canonical orientations from ShapeNet. To enable computing the occupancy, we also discard Breaking Bad objects that are not waterproof. After removal, we have 31,661 Breaking Bad objects.

\begin{figure*}[t]
    \centering
    \includegraphics[width=\linewidth]{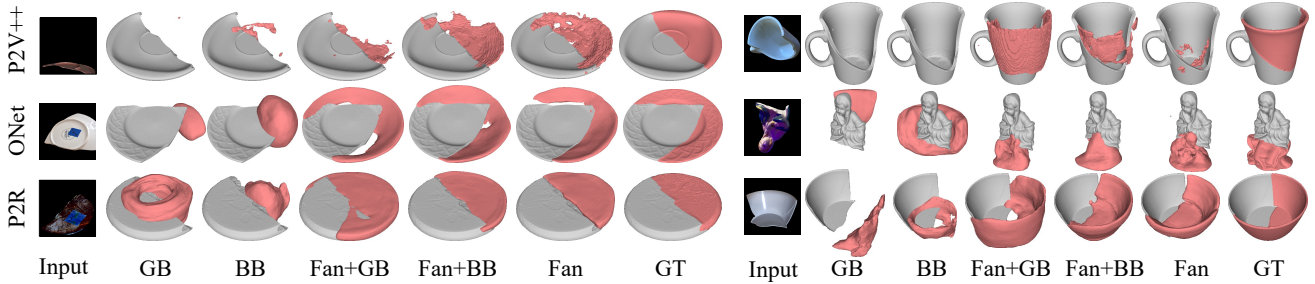}
    \caption{Repairs for images from our dataset for Pix2Vox++ (P2V++), ONet, and Pix2Repair (P2R) over various train sets and ground truth (GT). Fractured object in grey, repair in red.}
    \label{fig:rep}
\end{figure*}

We generate renders that look directly at the fractured region. We follow prior work that trains on synthetic renders~\cite{hui2022neural, chen2019learning, huang2022planes, mescheder2019occupancy}. To create renders, we compute the volumetric center of all vertices on the fractured region. We place a camera looking at the center of the fractured region vertices. We fit a plane to the fractured region vertices, and offset the camera in the direction of the plane normal. The normal may be pointing in the incorrect direction, i.e. pointing into the fractured object. To prevent incorrect normals, we compute the average dot product between the normal and the set of vectors pointing from the center of the fractured region vertices to the vertices in the fractured object, and flip the normal if the dot product is positive. We offset the camera in the direction of the normal such that it observes the entire object. We light the object using ambient lighting and using a camera-centered spotlight. We render objects at $224 \times 224$ pixel resolution using Pyrender~\cite{pyrender}.

We split each dataset into 70\%/30\% train/test sets within classes. As Geometric Breaks contains a high number of diverse objects (5,252) and a low number of fractures per object (1-10), we randomly select five renders from each fractured object in the dataset, giving us 18,170 training and 8,090 testing renders. As Breaking Bad contains a low number of diverse objects (405) and a high number of fractures per object (80), we randomly select one render from each fractured object, giving us 22,219 training and 10,181 testing renders. We use the same split as \textit{Fantastic Breaks} for our dataset, giving us 8,598 training and 3,039 testing images. We do not allow train/test object sets to overlap.

\setlength{\tabcolsep}{4pt}
\begin{table*}[t]
\centering
\footnotesize
\caption{Non-zero percent (NZ\%), chamfer distance (CD), and normal consistency (NC) for repair approaches tested on our dataset. Best values bolded.}
\begin{tabular}{l|ccc|ccc|ccc}
\toprule
\multicolumn{1}{c|}{} & \multicolumn{3}{c|}{Pix2Vox++} & \multicolumn{3}{c|}{ONet} & \multicolumn{3}{c}{Pix2Repair} \\
Train Set & NZ \% $\uparrow$ & CD $\downarrow$ & NC $\uparrow$ & NZ \% $\uparrow$ & CD $\downarrow$ & NC $\uparrow$ & NZ \% $\uparrow$ & CD $\downarrow$ & NC $\uparrow$ \\ \hline
GB & 8.2\% & 0.230 & 0.518 & 83.9\% & 0.165 & \textbf{0.520} & \textbf{95.5\%} & \textbf{0.125} & 0.497 \\
BB & 84.9\% & 0.147 & 0.530 & 87.0\% & 0.095 & 0.545 & \textbf{99.7\%} & \textbf{0.068} & \textbf{0.577} \\
BB + GB & 36.49\% & 0.193 & 0.497 & 87.4\% & 0.121 & 0.528 & \textbf{99.2\%} & \textbf{0.092} & \textbf{0.548} \\ \hline
Fan + GB & 83.1\% & 0.103 & 0.532 & 99.4\% & \textbf{0.062} & \textbf{0.635} & \textbf{99.9\%} & \textbf{0.062} & 0.599 \\
Fan + BB & 86.3\% & 0.108 & 0.533 & 99.5\% & \textbf{0.051} & 0.641 & \textbf{100.0\%} & \textbf{0.051} & \textbf{0.646} \\
Fan + BB + GB & 76.0\% & 0.134 & 0.538 & 99.7\% & 0.056 & \textbf{0.642} & \textbf{99.9\%} & \textbf{0.055} & 0.637 \\
Fan & 90.4\% & 0.087 & 0.532 & 99.9\% & \textbf{0.049} & \textbf{0.638} & \textbf{100.0\%} & 0.055 & 0.617 \\
\bottomrule
\end{tabular}
\label{tab:repairh}
\end{table*}

\subsection{Adapted Shape Repair Approaches} 

We adapt Onet and Pix2Vox++ to perform shape repair as our baselines. Each of the approaches takes images as input. ONet represents shapes implicitly using the occupancy function. Pix2Vox++ represents shapes as voxels. To adapt shape completion approaches for shape repair, we feed images of fractured objects as input and train each approach to predict the restoration shape. For Pix2Vox++ we implement their $128^3$ voxel resolution network, and train the network on $128^3$ resolution voxelized restoration shapes. For ONet, we use our point sampling approach to obtain $100,000$ uniform and surface sampled points.

\section{Results}\label{sec:res}


We show the quantitative results on our dataset in Table~\ref{tab:repairh} and show example restorations for objects from our dataset generated by Pix2Repair in Figure~\ref{fig:teaser}(h)~and~\ref{fig:rep}. The results demonstrate that Pix2Repair has the best performance in all but one case. They also illustrate that training repair approaches on the images of real fractured objects or synthetic datasets augmented with such real images is necessary to generate accurate restorations of real objects.

To evaluate predicted restoration shapes, we use the chamfer distance (CD) as defined by Park et al.~\cite{park2019deepsdf}. CD is an indicator of overall shape similarity, but may fail to capture out-of-distribution predictions~\cite{wu2021density}. A challenge of predicting canonically oriented restoration shapes is that the pose of the restoration shape cannot be determined for objects with rotational symmetry. As the location of the fractured region is independent of the object's canonical orientation, the network cannot accurately estimate where the restoration shape should be placed relative to the canonically oriented fractured shape. To evaluate symmetric shapes, we rotate the restoration shape around the up-axis $x$ times, and take the rotation that minimizes the chamfer distance between the predicted restoration shape and the ground truth shape to be the pose of the predicted restoration shape. We use $x=36$ for all classes in the \textit{Fantastic Breaks Imaged} dataset.
For success the CD should be low. 

We also evaluate shapes using the normal consistency (NC) as defined by Mescheder et al.~\cite{mescheder2019occupancy}. NC captures the alignment of surface normals between the predicted and ground truth shapes. For success, NC should be high. For all approaches we also show the non-zero percent (NZ\%). NZ\% denotes the percentage of fractured shapes for which a restoration shape is generated, i.e. for which the occupancy is not predicted as all zero. We compute CD and NC over 30,000 points.

We show the results of training on the images of synthetically fractured objects from Geometric Breaks (GB) and Breaking Bad (BB) and evaluating on \textit{Fantastic Breaks Imaged} in the first 3 rows of Table~\ref{tab:repairh}. We observe that all approaches struggle to generate accurate repair parts for objects when trained on synthetic data, as shown by the relatively high CD, e.g., of 0.125, 0.068, and 0.092, and 0.165, 0.095, and 0.121 for Pix2Repair and ONet when trained on GB, BB, and BB + GB respectively, compared to the bottom 4 rows of Table~\ref{tab:repairh}. Qualitative results in Figure~\ref{fig:rep} also show visually poor repairs for GB and BB. Pix2Vox++ struggles to repair a large percentage of objects, e.g., only 8.4\% when trained on GB, as small objects may be reduced to as few as 100 points during voxelization. We also observe networks trained on BB outperform networks trained on GB in terms of CD and NC, as shown in rows 1 and 2 of Table~\ref{tab:repairh}, likely as the physics-based fracture method of Breaking Bad is more realistic than the geometric method of Geometric Breaks.

The results of training on datasets that contain both synthetic and real fractured object images are shown in rows 4-7 of Table~\ref{tab:repairh}. For comparison, we also show results training only on objects from \textit{Fantastic Breaks Imaged} (Fan). We observe that adding real images to any combination of synthetic datasets increases performance, e.g., a CD of 0.125 to 0.062 and of 0.068 to 0.051 using Pix2Repair for GB and BB. We also observe that using \textit{Fantastic Breaks Imaged} without any synthetic data performs comparably to training using synthetic data with real data, e.g., with a CD of 0.051 with Fan+BB  and 0.055 for Fan using Pix2Repair, and with a CD of 0.051 with Fan+BB and 0.049 with Fan using ONet, indicating that for some objects our image dataset contains sufficient variation to generalize to new 3D shapes. Figure~\ref{fig:rep} shows that most approaches trained on real data produce visually appealing repair parts, e.g., the plate for Onet and the coaster and bowl for Pix2Repair. Similar to training on synthetic datasets only, Pix2Repair outperforms shape completion approaches adapted for completion, except for ONet on Fan.  We also observe that combining Breaking Bad with our dataset produces some of the best repairs in terms of CD and NC, as shown in row 5 of Table~\ref{tab:repairh} and column 5 of Figure~\ref{fig:rep}, e.g., with a CD of 0.051 for Onet and NC of 0.646 for Pix2Repair.

\section{Conclusion}
\label{sec:con}

\begin{figure}
\centering
\includegraphics[width=.7\linewidth]{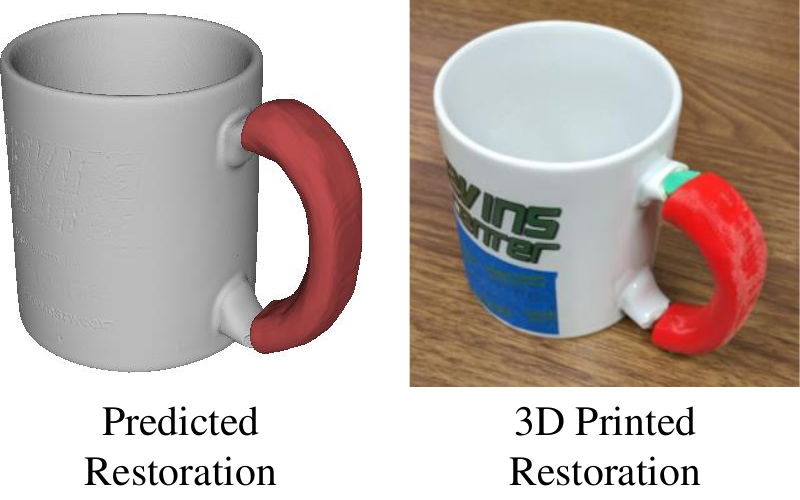}
\caption{3D printed repairs must be re-scaled.}
\label{fig:limit}
\end{figure}

We present Pix2Repair, an approach to automatically generate restoration shapes from images. Our approach outperforms existing shape completion approaches adapted for shape repair. We also introduce \textit{Fantastic Breaks Imaged}, the first large-scale dataset containing 11,637 images corresponding to models from \textit{Fantastic Breaks} complete with rich annotations for facilitating future shape repair research. We train Pix2Repai on the \textit{Fantastic Breaks Imaged} dataset, synthetic dataset, and mixed datasets, and evaluate it on \textit{Fantastic Breaks Imaged}. In contrast to existing approaches that require high-resolution 3D models, by operating on images of fractured objects Pix2Repair represents a significant step forward in increasing the accessibility of repair algorithms. Our approach and dataset also have some limitations. For example, a single image of the fractured object may not observe parts of the object that are essential to determining its overall structure. Another limitation is that restoration shapes predicted by our approach are not accurate to scale, preventing them from being used directly for 3D printing. It is not possible to accurately determine object scale from color images without external information. We manually re-scale, 3D print, and attach a predicted restoration shape to a fractured object with putty in Figure~\ref{fig:limit}. Re-scaling may inhibit the restoration part from fitting snugly. As part of future work, we will explore predicting restoration shapes from other modalities that include scale information, i.e. depth images.


{
    \small
    \bibliographystyle{ieeenat_fullname}
    \bibliography{main}

\begin{thebibliography}{63}
\providecommand{\natexlab}[1]{#1}
\providecommand{\url}[1]{\texttt{#1}}
\expandafter\ifx\csname urlstyle\endcsname\relax
  \providecommand{\doi}[1]{doi: #1}\else
  \providecommand{\doi}{doi: \begingroup \urlstyle{rm}\Url}\fi

\bibitem[Antlej et~al.(2011)Antlej, Eric, Savnik, Zupanek, Slabe, and Battestin]{antlej2011combining}
Kaja Antlej, Miran Eric, Mojca Savnik, Bernarda Zupanek, Janja Slabe, and B~Borut Battestin.
\newblock Combining 3d technologies in the field of cultural heritage: Three case studies.
\newblock In \emph{VAST (Short and Project Papers)}, 2011.

\bibitem[Bekisz et~al.(2019)Bekisz, Liss, Maliha, Witek, Coelho, and Flores]{bekisz2019house}
Jonathan~M Bekisz, Hannah~A Liss, Samantha~G Maliha, Lukasz Witek, Paulo~G Coelho, and Roberto~L Flores.
\newblock In-house manufacture of sterilizable, scaled, patient-specific 3d-printed models for rhinoplasty.
\newblock \emph{Aesthetic Surgery Journal}, 39\penalty0 (3):\penalty0 254--263, 2019.

\bibitem[Brown et~al.(2008)Brown, Toler-Franklin, Nehab, Burns, Dobkin, Vlachopoulos, Doumas, Rusinkiewicz, and Weyrich]{brown2008system}
Benedict~J Brown, Corey Toler-Franklin, Diego Nehab, Michael Burns, David Dobkin, Andreas Vlachopoulos, Christos Doumas, Szymon Rusinkiewicz, and Tim Weyrich.
\newblock A system for high-volume acquisition and matching of fresco fragments: Reassembling theran wall paintings.
\newblock \emph{ACM transactions on graphics (TOG)}, 27\penalty0 (3):\penalty0 1--9, 2008.

\bibitem[Chang et~al.(2015)Chang, Funkhouser, Guibas, Hanrahan, Huang, Li, Savarese, Savva, Song, Su, Xiao, Yi, and Yu]{shapenet2015}
Angel~X. Chang, Thomas Funkhouser, Leonidas Guibas, Pat Hanrahan, Qixing Huang, Zimo Li, Silvio Savarese, Manolis Savva, Shuran Song, Hao Su, Jianxiong Xiao, Li Yi, and Fisher Yu.
\newblock {ShapeNet: An Information-Rich 3D Model Repository}.
\newblock Technical Report arXiv:1512.03012 [cs.GR], Stanford University --- Princeton University --- Toyota Technological Institute at Chicago, 2015.

\bibitem[Chen et~al.(2022)Chen, Li, Turpin, Jacobson, and Garg]{chen2022neural}
Yun-Chun Chen, Haoda Li, Dylan Turpin, Alec Jacobson, and Animesh Garg.
\newblock Neural shape mating: Self-supervised object assembly with adversarial shape priors.
\newblock In \emph{Proceedings of the IEEE/CVF Conference on Computer Vision and Pattern Recognition}, pages 12724--12733, 2022.

\bibitem[Chen and Zhang(2019)]{chen2019learning}
Zhiqin Chen and Hao Zhang.
\newblock Learning implicit fields for generative shape modeling.
\newblock In \emph{Proc. CVPR}, pages 5939--5948, Piscataway, NJ, 2019. IEEE.

\bibitem[Choy et~al.(2016)Choy, Xu, Gwak, Chen, and Savarese]{choy20163d}
Christopher~B Choy, Danfei Xu, JunYoung Gwak, Kevin Chen, and Silvio Savarese.
\newblock 3d-r2n2: A unified approach for single and multi-view 3d object reconstruction.
\newblock In \emph{European conference on computer vision}, pages 628--644. Springer, 2016.

\bibitem[Dai et~al.(2020)Dai, Diller, and Nie{\ss}ner]{dai2020sg}
Angela Dai, Christian Diller, and Matthias Nie{\ss}ner.
\newblock Sg-nn: Sparse generative neural networks for self-supervised scene completion of rgb-d scans.
\newblock In \emph{Proceedings of the IEEE/CVF Conference on Computer Vision and Pattern Recognition}, pages 849--858, 2020.

\bibitem[Downs et~al.(2022)Downs, Francis, Koenig, Kinman, Hickman, Reymann, McHugh, and Vanhoucke]{downs2022google}
Laura Downs, Anthony Francis, Nate Koenig, Brandon Kinman, Ryan Hickman, Krista Reymann, Thomas~B McHugh, and Vincent Vanhoucke.
\newblock Google scanned objects: A high-quality dataset of 3d scanned household items.
\newblock In \emph{2022 International Conference on Robotics and Automation (ICRA)}, pages 2553--2560. IEEE, 2022.

\bibitem[Funkhouser et~al.(2011)Funkhouser, Shin, Toler-Franklin, Casta{\~n}eda, Brown, Dobkin, Rusinkiewicz, and Weyrich]{funkhouser2011learning}
Thomas Funkhouser, Hijung Shin, Corey Toler-Franklin, Antonio~Garc{\'\i}a Casta{\~n}eda, Benedict Brown, David Dobkin, Szymon Rusinkiewicz, and Tim Weyrich.
\newblock Learning how to match fresco fragments.
\newblock \emph{Journal on Computing and Cultural Heritage (JOCCH)}, 4\penalty0 (2):\penalty0 1--13, 2011.

\bibitem[Gkioxari et~al.(2019)Gkioxari, Malik, and Johnson]{gkioxari2019mesh}
Georgia Gkioxari, Jitendra Malik, and Justin Johnson.
\newblock Mesh r-cnn.
\newblock In \emph{Proceedings of the IEEE/CVF International Conference on Computer Vision}, pages 9785--9795, 2019.

\bibitem[Gondal et~al.(2019)Gondal, Wuthrich, Miladinovic, Locatello, Breidt, Volchkov, Akpo, Bachem, Sch{\"o}lkopf, and Bauer]{gondal2019transfer}
Muhammad~Waleed Gondal, Manuel Wuthrich, Djordje Miladinovic, Francesco Locatello, Martin Breidt, Valentin Volchkov, Joel Akpo, Olivier Bachem, Bernhard Sch{\"o}lkopf, and Stefan Bauer.
\newblock On the transfer of inductive bias from simulation to the real world: a new disentanglement dataset.
\newblock \emph{Advances in Neural Information Processing Systems}, 32, 2019.

\bibitem[Groueix et~al.(2018)Groueix, Fisher, Kim, Russell, and Aubry]{groueix2018papier}
Thibault Groueix, Matthew Fisher, Vladimir~G Kim, Bryan~C Russell, and Mathieu Aubry.
\newblock A papier-m{\^a}ch{\'e} approach to learning 3d surface generation.
\newblock In \emph{Proc. CVPR}, pages 216--224, Piscataway, NJ, 2018. IEEE.

\bibitem[Gupta and Qi(1991)]{gupta1991theory}
Madan~M Gupta and J11043360726 Qi.
\newblock Theory of t-norms and fuzzy inference methods.
\newblock \emph{Fuzzy sets and systems}, 40\penalty0 (3):\penalty0 431--450, 1991.

\bibitem[Hermoza and Sipiran(2018)]{hermoza20183d}
Renato Hermoza and Ivan Sipiran.
\newblock 3d reconstruction of incomplete archaeological objects using a generative adversarial network.
\newblock In \emph{Proceedings of Computer Graphics International}, pages 5--11. ACM, New York, NY, 2018.

\bibitem[Hong et~al.(2019)Hong, Kim, Wi, and Kim]{hong2019potsac}
Je~Hyeong Hong, Young~Min Kim, Koang-Chul Wi, and Jinwook Kim.
\newblock Potsac: A robust axis estimator for axially symmetric pot fragments.
\newblock In \emph{ICCV Workshops}, pages 1421--1428, 2019.

\bibitem[Huang et~al.(2006)Huang, Fl{\"o}ry, Gelfand, Hofer, and Pottmann]{huang2006reassembling}
Qi-Xing Huang, Simon Fl{\"o}ry, Natasha Gelfand, Michael Hofer, and Helmut Pottmann.
\newblock Reassembling fractured objects by geometric matching.
\newblock In \emph{ACM SIGGRAPH}, pages 569--578, 2006.

\bibitem[Huang et~al.(2022)Huang, Stojanov, Thai, Jampani, and Rehg]{huang2022planes}
Zixuan Huang, Stefan Stojanov, Anh Thai, Varun Jampani, and James~M Rehg.
\newblock Planes vs. chairs: Category-guided 3d shape learning without any 3d cues.
\newblock In \emph{European Conference on Computer Vision}, pages 727--744. Springer, 2022.

\bibitem[Hui et~al.(2022)Hui, Li, Hu, and Fu]{hui2022neural}
Ka-Hei Hui, Ruihui Li, Jingyu Hu, and Chi-Wing Fu.
\newblock Neural template: Topology-aware reconstruction and disentangled generation of 3d meshes.
\newblock In \emph{Proceedings of the IEEE/CVF Conference on Computer Vision and Pattern Recognition}, pages 18572--18582, 2022.

\bibitem[Ioffe and Szegedy(2015)]{ioffe2015batch}
Sergey Ioffe and Christian Szegedy.
\newblock Batch normalization: Accelerating deep network training by reducing internal covariate shift.
\newblock In \emph{International conference on machine learning}, pages 448--456. pmlr, 2015.

\bibitem[Kingma and Ba(2014)]{kingma2014adam}
Diederik~P Kingma and Jimmy Ba.
\newblock Adam: A method for stochastic optimization.
\newblock In \emph{Proc. ICLR}, pages 1--15, La Jolla, CA, 2014. International Conference on Representation Learning.

\bibitem[Kirillov et~al.(2023)Kirillov, Mintun, Ravi, Mao, Rolland, Gustafson, Xiao, Whitehead, Berg, Lo, et~al.]{kirillov2023segment}
Alexander Kirillov, Eric Mintun, Nikhila Ravi, Hanzi Mao, Chloe Rolland, Laura Gustafson, Tete Xiao, Spencer Whitehead, Alexander~C Berg, Wan-Yen Lo, et~al.
\newblock Segment anything.
\newblock \emph{arXiv preprint arXiv:2304.02643}, 2023.

\bibitem[Lamb et~al.(2019)Lamb, Banerjee, and Banerjee]{lamb2019automated}
Nikolas Lamb, Sean Banerjee, and Natasha~Kholgade Banerjee.
\newblock Automated reconstruction of smoothly joining 3d printed restorations to fix broken objects.
\newblock In \emph{Proc. SCF}, pages 1--12, New York, NY, 2019. ACM.

\bibitem[Lamb et~al.(2022{\natexlab{a}})Lamb, Banerjee, and Banerjee]{lamb2022deepjoin}
Nikolas Lamb, Sean Banerjee, and Natasha~Kholgade Banerjee.
\newblock Deepjoin: Learning a joint occupancy, signed distance, and normal field function for shape repair.
\newblock \emph{ACM Transactions on Graphics (TOG)}, 41\penalty0 (6):\penalty0 1--10, 2022{\natexlab{a}}.

\bibitem[Lamb et~al.(2022{\natexlab{b}})Lamb, Banerjee, and Banerjee]{lamb2022deepmend}
Nikolas Lamb, Sean Banerjee, and Natasha~Kholgade Banerjee.
\newblock Deepmend: Learning occupancy functions to represent shape for repair.
\newblock In \emph{European Conference on Computer Vision}, pages 433--450. Springer, 2022{\natexlab{b}}.

\bibitem[Lamb et~al.(2022{\natexlab{c}})Lamb, Banerjee, and Banerjee]{lamb2022mendnet}
Nikolas Lamb, Sean Banerjee, and Natasha~K. Banerjee.
\newblock {MendNet: Restoration of Fractured Shapes Using Learned Occupancy Functions}.
\newblock \emph{Computer Graphics Forum}, 2022{\natexlab{c}}.

\bibitem[Lamb et~al.(2023)Lamb, Palmer, Molloy, Banerjee, and Banerjee]{lamb2023fantastic}
Nikolas Lamb, Cameron Palmer, Ben Molloy, Sean Banerjee, and Natasha~Kholgade Banerjee.
\newblock Fantastic breaks: A dataset of paired 3d scans of real-world broken objects and their complete counterparts.
\newblock In \emph{Proceedings of the IEEE conference on computer vision and pattern recognition}, 2023.

\bibitem[Li et~al.(2014)Li, Zhang, and Chen]{li2014symmetry}
Er Li, Xiaopeng Zhang, and Yanyun Chen.
\newblock Symmetry based chinese ancient architecture reconstruction from incomplete point cloud.
\newblock In \emph{2014 5th International conference on digital home}, pages 157--161. IEEE, 2014.

\bibitem[Li and Zhang(2021)]{li2021d2im}
Manyi Li and Hao Zhang.
\newblock D2im-net: Learning detail disentangled implicit fields from single images.
\newblock In \emph{Proceedings of the IEEE/CVF Conference on Computer Vision and Pattern Recognition}, pages 10246--10255, 2021.

\bibitem[Lin et~al.(2020)Lin, Wang, and Lucey]{lin2020sdf}
Chen-Hsuan Lin, Chaoyang Wang, and Simon Lucey.
\newblock Sdf-srn: Learning signed distance 3d object reconstruction from static images.
\newblock \emph{arXiv preprint arXiv:2010.10505}, 1\penalty0 (1):\penalty0 1--17, 2020.

\bibitem[Lorensen and Cline(1987)]{lorensen1987marching}
William~E Lorensen and Harvey~E Cline.
\newblock Marching cubes: A high resolution 3d surface construction algorithm.
\newblock \emph{ACM SIGGRAPH Computer Graphics}, 21\penalty0 (4):\penalty0 163--169, 1987.

\bibitem[Ma and Wang(2023)]{Ma2023SegmentAI}
Jun Ma and Bo Wang.
\newblock Segment anything in medical images.
\newblock \emph{ArXiv}, abs/2304.12306, 2023.

\bibitem[Mandikal et~al.(2018)Mandikal, Navaneet, Agarwal, and Babu]{mandikal20183d}
Priyanka Mandikal, KL Navaneet, Mayank Agarwal, and R~Venkatesh Babu.
\newblock 3d-lmnet: Latent embedding matching for accurate and diverse 3d point cloud reconstruction from a single image.
\newblock \emph{arXiv preprint arXiv:1807.07796}, 2018.

\bibitem[Matl(2023)]{pyrender}
Matthew Matl.
\newblock Pyrender, 2023.

\bibitem[Mavridis et~al.(2015)Mavridis, Sipiran, Andreadis, and Papaioannou]{mavridis2015object}
Pavlos Mavridis, Ivan Sipiran, Anthousis Andreadis, and Georgios Papaioannou.
\newblock Object completion using k-sparse optimization.
\newblock In \emph{Computer Graphics Forum}, pages 13--21. Wiley Online Library, 2015.

\bibitem[Mescheder et~al.(2019)Mescheder, Oechsle, Niemeyer, Nowozin, and Geiger]{mescheder2019occupancy}
Lars Mescheder, Michael Oechsle, Michael Niemeyer, Sebastian Nowozin, and Andreas Geiger.
\newblock Occupancy networks: Learning 3d reconstruction in function space.
\newblock In \emph{Proc. CVPR}, pages 4460--4470, Piscataway, NJ, 2019. IEEE.

\bibitem[Mo et~al.(2019)Mo, Zhu, Chang, Yi, Tripathi, Guibas, and Su]{mo2019partnet}
Kaichun Mo, Shilin Zhu, Angel~X Chang, Li Yi, Subarna Tripathi, Leonidas~J Guibas, and Hao Su.
\newblock Partnet: A large-scale benchmark for fine-grained and hierarchical part-level 3d object understanding.
\newblock In \emph{Proceedings of the IEEE/CVF conference on computer vision and pattern recognition}, pages 909--918, 2019.

\bibitem[Murez et~al.(2020)Murez, Van~As, Bartolozzi, Sinha, Badrinarayanan, and Rabinovich]{murez2020atlas}
Zak Murez, Tarrence Van~As, James Bartolozzi, Ayan Sinha, Vijay Badrinarayanan, and Andrew Rabinovich.
\newblock Atlas: End-to-end 3d scene reconstruction from posed images.
\newblock In \emph{European conference on computer vision}, pages 414--431. Springer, 2020.

\bibitem[Park et~al.(2019)Park, Florence, Straub, Newcombe, and Lovegrove]{park2019deepsdf}
Jeong~Joon Park, Peter Florence, Julian Straub, Richard Newcombe, and Steven Lovegrove.
\newblock Deepsdf: Learning continuous signed distance functions for shape representation.
\newblock In \emph{Proc. CVPR}, pages 165--174, Piscataway, NJ, 2019. IEEE.

\bibitem[Payne et~al.(2009)Payne, Cole, Simon, Goodmaster, and Limp]{payne2009designing}
Angelia Payne, Keenan Cole, Katie Simon, Christopher Goodmaster, and Fredrick Limp.
\newblock Designing the next generation virtual museum: Making 3d artifacts available for viewing and download.
\newblock In \emph{Making History Interactive: Proceedings of the 37th Annual International Conference on Computer Applications and Quantitative Methods in Archaeology (CAA)}, pages 1--6, 2009.

\bibitem[Ping et~al.(2022)Ping, Esfahani, Chen, and Wang]{ping2022visual}
Guiju Ping, Mahdi~Abolfazli Esfahani, Jiaying Chen, and Han Wang.
\newblock Visual enhancement of single-view 3d point cloud reconstruction.
\newblock \emph{Computers \& Graphics}, 102:\penalty0 112--119, 2022.

\bibitem[Popov et~al.(2020)Popov, Bauszat, and Ferrari]{popov2020corenet}
Stefan Popov, Pablo Bauszat, and Vittorio Ferrari.
\newblock Corenet: Coherent 3d scene reconstruction from a single rgb image.
\newblock In \emph{European Conference on Computer Vision}, pages 366--383. Springer, 2020.

\bibitem[Rengier et~al.(2010)Rengier, Mehndiratta, Von Tengg-Kobligk, Zechmann, Unterhinninghofen, Kauczor, and Giesel]{rengier20103d}
Fabian Rengier, Amit Mehndiratta, Hendrik Von Tengg-Kobligk, Christian~M Zechmann, Roland Unterhinninghofen, H-U Kauczor, and Frederik~L Giesel.
\newblock 3d printing based on imaging data: review of medical applications.
\newblock \emph{International journal of computer assisted radiology and surgery}, 5\penalty0 (4):\penalty0 335--341, 2010.

\bibitem[Schilling et~al.(2014)Schilling, Jastram, Wings, Schwarz-Wings, and Issever]{schilling2014reviving}
Ren{\'e} Schilling, Benjamin Jastram, Oliver Wings, Daniela Schwarz-Wings, and Ahi~Sema Issever.
\newblock Reviving the dinosaur: virtual reconstruction and three-dimensional printing of a dinosaur vertebra.
\newblock \emph{Radiology}, 270\penalty0 (3):\penalty0 864--871, 2014.

\bibitem[Scopigno et~al.(2011)Scopigno, Callieri, Cignoni, Corsini, Dellepiane, Ponchio, and Ranzuglia]{scopigno20113d}
Roberto Scopigno, Marco Callieri, Paolo Cignoni, Massimiliano Corsini, Matteo Dellepiane, Federico Ponchio, and Guido Ranzuglia.
\newblock 3d models for cultural heritage: Beyond plain visualization.
\newblock \emph{Computer}, 44\penalty0 (7):\penalty0 48--55, 2011.

\bibitem[Seixas et~al.(2018)Seixas, Assis, Figueiredo, Pinto, and Paula]{seixas2018use}
Maria~Luiza Seixas, Paulo~Santos Assis, Jo{\~a}o Cura~D’Ars Figueiredo, Maria~Aparecida Pinto, and Daniella Gualberto~Caldeira Paula.
\newblock The use of rapid prototyping in the joining of fractured historical silver object.
\newblock \emph{Rapid Prototyping Journal}, 2018.

\bibitem[Sell{\'a}n et~al.(2022)Sell{\'a}n, Chen, Wu, Garg, and Jacobson]{sellan2022breaking}
S. Sell{\'a}n, Y. Chen, Z. Wu, A. Garg, and A. Jacobson.
\newblock Breaking bad: A dataset for geometric fracture and reassembly.
\newblock In \emph{Thirty-sixth Conference on Neural Information Processing Systems Datasets and Benchmarks Track}, 2022.

\bibitem[Shi et~al.(2021)Shi, Ni, Liu, Rong, Qian, and Zhang]{shi2021geometric}
Yue Shi, Bingbing Ni, Jinxian Liu, Dingyi Rong, Ye Qian, and Wenjun Zhang.
\newblock Geometric granularity aware pixel-to-mesh.
\newblock In \emph{Proceedings of the IEEE/CVF International Conference on Computer Vision}, pages 13097--13106, 2021.

\bibitem[Sipiran(2018)]{sipiran2018completion}
Ivan Sipiran.
\newblock Completion of cultural heritage objects with rotational symmetry.
\newblock In \emph{Proceedings of the 11th Eurographics Workshop on 3D Object Retrieval}, pages 87--93, 2018.

\bibitem[Wang et~al.(2018)Wang, Zhang, Li, Fu, Liu, and Jiang]{wang2018pixel2mesh}
Nanyang Wang, Yinda Zhang, Zhuwen Li, Yanwei Fu, Wei Liu, and Yu-Gang Jiang.
\newblock Pixel2mesh: Generating 3d mesh models from single rgb images.
\newblock In \emph{Proceedings of the European conference on computer vision (ECCV)}, pages 52--67, 2018.

\bibitem[Wang et~al.(2019)Wang, Ceylan, Mech, and Neumann]{wang20193dn}
Weiyue Wang, Duygu Ceylan, Radomir Mech, and Ulrich Neumann.
\newblock 3dn: 3d deformation network.
\newblock In \emph{Proceedings of the IEEE/CVF Conference on Computer Vision and Pattern Recognition}, pages 1038--1046, 2019.

\bibitem[Wang et~al.(2022)Wang, Zhuang, Liu, and Chen]{wang2022mdisn}
Yujie Wang, Yixin Zhuang, Yunzhe Liu, and Baoquan Chen.
\newblock Mdisn: Learning multiscale deformed implicit fields from single images.
\newblock \emph{Visual Informatics}, 2022.

\bibitem[Wu et~al.(2017)Wu, Wang, Xue, Sun, Freeman, and Tenenbaum]{wu2017marrnet}
Jiajun Wu, Yifan Wang, Tianfan Xue, Xingyuan Sun, Bill Freeman, and Josh Tenenbaum.
\newblock Marrnet: 3d shape reconstruction via 2.5 d sketches.
\newblock \emph{Advances in neural information processing systems}, 30, 2017.

\bibitem[Wu et~al.(2018)Wu, Zhang, Zhang, Zhang, Freeman, and Tenenbaum]{wu2018learning}
Jiajun Wu, Chengkai Zhang, Xiuming Zhang, Zhoutong Zhang, William~T Freeman, and Joshua~B Tenenbaum.
\newblock Learning shape priors for single-view 3d completion and reconstruction.
\newblock In \emph{Proceedings of the European Conference on Computer Vision (ECCV)}, pages 646--662, 2018.

\bibitem[Wu et~al.(2020)Wu, Zhuang, Xu, Zhang, and Chen]{wu2020pq}
Rundi Wu, Yixin Zhuang, Kai Xu, Hao Zhang, and Baoquan Chen.
\newblock Pq-net: A generative part seq2seq network for 3d shapes.
\newblock In \emph{Proceedings of the IEEE/CVF Conference on Computer Vision and Pattern Recognition}, pages 829--838, 2020.

\bibitem[Wu et~al.(2021)Wu, Pan, Zhang, Wang, Liu, and Lin]{wu2021density}
Tong Wu, Liang Pan, Junzhe Zhang, Tai Wang, Ziwei Liu, and Dahua Lin.
\newblock Density-aware chamfer distance as a comprehensive metric for point cloud completion.
\newblock \emph{arXiv preprint arXiv:2111.12702}, 2021.

\bibitem[Wu et~al.(2022)Wu, Wang, Pan, Xu, Theobalt, Liu, and Lin]{wu2022voxurf}
Tong Wu, Jiaqi Wang, Xingang Pan, Xudong Xu, Christian Theobalt, Ziwei Liu, and Dahua Lin.
\newblock Voxurf: Voxel-based efficient and accurate neural surface reconstruction.
\newblock \emph{arXiv preprint arXiv:2208.12697}, 2022.

\bibitem[Xie et~al.(2019)Xie, Yao, Sun, Zhou, and Zhang]{xie2019pix2vox}
Haozhe Xie, Hongxun Yao, Xiaoshuai Sun, Shangchen Zhou, and Shengping Zhang.
\newblock Pix2vox: Context-aware 3d reconstruction from single and multi-view images.
\newblock In \emph{Proceedings of the IEEE/CVF international conference on computer vision}, pages 2690--2698, 2019.

\bibitem[Xie et~al.(2020)Xie, Yao, Zhang, Zhou, and Sun]{xie2020pix2vox++}
Haozhe Xie, Hongxun Yao, Shengping Zhang, Shangchen Zhou, and Wenxiu Sun.
\newblock Pix2vox++: Multi-scale context-aware 3d object reconstruction from single and multiple images.
\newblock \emph{International Journal of Computer Vision}, 128\penalty0 (12):\penalty0 2919--2935, 2020.

\bibitem[Xu et~al.(2020)Xu, Fan, Yuan, and Singh]{xu2020ladybird}
Yifan Xu, Tianqi Fan, Yi Yuan, and Gurprit Singh.
\newblock Ladybird: Quasi-monte carlo sampling for deep implicit field based 3d reconstruction with symmetry.
\newblock In \emph{European Conference on Computer Vision}, pages 248--263. Springer, 2020.

\bibitem[Yi et~al.(2021)Yi, Gong, and Funkhouser]{yi2021complete}
Li Yi, Boqing Gong, and Thomas Funkhouser.
\newblock Complete \& label: A domain adaptation approach to semantic segmentation of lidar point clouds.
\newblock In \emph{Proceedings of the IEEE/CVF Conference on Computer Vision and Pattern Recognition}, pages 15363--15373, 2021.

\bibitem[Yu et~al.(2022)Yu, Yang, and Wei]{yu2022part}
Qian Yu, Chengzhuan Yang, and Hui Wei.
\newblock Part-wise atlasnet for 3d point cloud reconstruction from a single image.
\newblock \emph{Knowledge-Based Systems}, page 108395, 2022.

\bibitem[Zhou and Jacobson(2016)]{zhou2016thingi10k}
Qingnan Zhou and Alec Jacobson.
\newblock Thingi10k: A dataset of 10,000 3d-printing models.
\newblock \emph{arXiv preprint arXiv:1605.04797}, 2016.

\end{thebibliography}
}


\end{document}